\documentclass[conference]{IEEEtran}
\usepackage{cite}
\usepackage{amsmath,amssymb,amsfonts}
\usepackage{graphicx}
\usepackage{booktabs}
\usepackage{multirow}
\usepackage{xcolor}
\usepackage[hidelinks]{hyperref}
\usepackage{balance}

\begin{document}

\title{Explainable AI for Screening Abuse-Related Trauma in Bangladeshi Children: A Training-Free Multimodal Framework Evaluated on Noise-Aware Synthetic Data}

\author{
\IEEEauthorblockN{Salma Hoque Talukdar Koli}
\IEEEauthorblockA{\textit{Dept.\ of Computer Science \& Engineering}\\
\textit{RTM Al-Kabir Technical University}\\
Sylhet-3100, Bangladesh\\
info.salmahoquetalukdarkoli@gmail.com}
\and
\IEEEauthorblockN{Fahima Haque Talukder Jely}
\IEEEauthorblockA{\textit{Dept.\ of Computer Science \& Engineering}\\
\textit{North East University Bangladesh}\\
Sylhet, Bangladesh\\
fahimahaquetalukderjely@gmail.com}
}

\maketitle

\begin{abstract}
Bangladesh has an estimated 1.17 mental-health professionals per
100{,}000 population and only six child psychiatrists nationwide. No
Bengali-language, culturally adapted tool exists for early screening of
abuse-related psychological trauma in children. We present
\emph{ShishuRaksha AI}, a decision-support (not diagnostic) framework
that fuses four screening modalities: validated questionnaires (SDQ,
CPSS), Bengali narrative text, House-Tree-Person (HTP) drawing
features, and facial affect. The fusion is training-free and clinically
weighted, uses cross-modal attention, and includes a single-modality
override rule. Every risk score is explained through clinically
weighted, perturbation-based additive attribution and rendered as a
bilingual (Bangla/English) report with referral routing to national
child-protection services (OCC, DSS, NMHH) under the Children Act 2013.
No clinical dataset of abused children can be collected ethically at
this stage, so we introduce a noise-aware synthetic benchmark (500
cases, 116 positive [23.2\%], four deliberate noise layers,
literature-grounded HTP priors) and evaluate tree-ensemble surrogates
of the fusion design (facial channel excluded) under 5-fold stratified
cross-validation. The fused model reaches an AUC of 0.874
[0.834--0.908], against 0.756 [0.705--0.803] for an SDQ-only baseline,
with ablation, operating-point, subgroup, and calibration analyses. We
state all limitations openly, including synthetic-only data, no
held-out set, text-feature circularity, and an urban--rural subgroup
gap. This work is a feasibility study and a design contribution toward
ethically deployable child-protection screening in low-resource
settings.
\end{abstract}


\section{Introduction}
A child in a remote haor district such as Sunamganj who stops speaking after a violent episode at
home will, in most cases, never see a specialist. Bangladesh has an
estimated 1.17 mental-health professionals per 100{,}000 population and,
per WHO figures, only six child psychiatrists in the entire
country~\cite{koly2024stakeholder}. The epidemiology is not reassuring: community surveys using the Bangla SDQ have found
psychiatric disorder in substantial fractions of 5--10-year-olds across
rural, urban, and slum settings alike~\cite{mullick2005prevalence}.
Abuse-related trauma sits in the worst corner of this gap. It is
stigmatized and rarely disclosed, and outside a handful of urban
clinics it is almost never screened. The instruments that do exist are
paper questionnaires that need trained administrators, written in
English or translated without any automation.

Machine learning cannot close this gap, and we do not claim it can.
What it might do is triage: help a school counselor, an NGO field
worker, or a shishu welfare officer decide \emph{which} children need
referral to the scarce professionals who exist. That framing
shapes every design choice here. ShishuRaksha AI\footnote{Code and the full synthetic benchmark: \url{https://github.com/shkoli/shishuraksha-ai}} produces a
prioritization signal with a human-readable justification, in Bangla,
routed to Bangladesh's real child-protection bodies: One-Stop Crisis
Centres (OCC), the Department of Social Services (DSS), and the
national mental-health helpline, under the Children Act
2013~\cite{childrenact2013}. It is decision support; it does not
diagnose, and it is built so that it cannot quietly pretend to.

There is an obvious ethical problem with building such a system:
training or validating on records of abused children cannot be
justified at the proof-of-concept stage, and no ethics board was asked
to permit it. We take a synthetic-first path instead. A noise-aware
generator produces a benchmark cohort grounded in published HTP and
screening literature, and the framework is evaluated on it with the
full set of checks a real validation would require: ablation,
operating points, subgroups, and calibration. The numbers are feasibility evidence, nothing
more; Section~\ref{sec:limitations} is deliberately the longest
discussion in the paper.

The contributions: a Bangladesh-specific four-modality screening
framework; training-free attention fusion with clinically specified
weights and a single-modality override; bilingual explainable
reporting wired to statutory referral routing; a noise-aware synthetic
benchmark methodology for domains where real data collection is
ethically infeasible; and a feasibility evaluation of the whole
design.

\section{Related Work}

\subsection{Child Psychological Screening Instruments}
The Strengths and Difficulties Questionnaire (SDQ)~\cite{goodman1997sdq}
is the most widely deployed brief screening instrument for child mental
health, with well-established psychometric
properties~\cite{goodman2001psychometric}. A Bangla adaptation was
validated by Mullick and Goodman on 261 Bangladeshi children,
distinguishing clinic from community samples and between diagnostic
groups~\cite{mullick2001bangla}, the SDQ multi-informant algorithm has
been evaluated in Dhaka clinics alongside London~\cite{goodman2000dhaka},
and Bangla-SDQ epidemiology has documented substantial prevalence of
child psychiatric disorder across rural, urban, and slum communities in
Bangladesh~\cite{mullick2005prevalence}. For trauma specifically, the
Child PTSD Symptom Scale (CPSS)~\cite{foa2001cpss} is a standard
self-report measure. All of these instruments, however, assume access
to trained administrators and, in Bangladesh, confront a system with
only six child psychiatrists nationwide~\cite{koly2024stakeholder};
none provides automated triage support in Bengali.

\subsection{Projective Drawing Analysis}
The House-Tree-Person (HTP) test~\cite{buck1948htp} and related drawing
techniques~\cite{malchiodi1998art} have long been used with children
precisely because they demand little verbal ability. The evidence behind them is mixed: Allen and Tussey's systematic review found that
projective drawings could not reliably detect whether a child had
experienced sexual or physical abuse~\cite{allen2012projective}, and
Lin et al.\ showed that deep neural networks trained on 4{,}196
children's HTP drawings failed to predict depression, casting doubt on
HTP as a stand-alone diagnostic signal~\cite{lin2022htp}. At the same
time, computational drawing analysis continues to advance:
object-detection pipelines now generate structured HTP score tables
automatically~\cite{lee2024htp}, machine learning over digitized
drawings has been used to identify psychological trauma among Syrian
refugee children for early intervention~\cite{baird2022syrian}, and
recent computer-vision systems automate feature extraction from
children's drawings for screening referral~\cite{alshahrani2024drawing}.
We take the adverse validity evidence seriously as a design constraint
rather than ignoring it: in ShishuRaksha AI the drawing modality is
(i) one weak signal among four, (ii) down-weighted by clinically
specified weights, and (iii) used only for \emph{screening
prioritization}, never diagnosis.

\subsection{Multimodal Machine Learning for Mental Health}
Machine learning has been applied across mental-health detection,
diagnosis support, and public-health surveillance~\cite{shatte2019ml},
and multimodal approaches (for example, fusing audio and text of
clinical interviews for depression
detection~\cite{alhanai2018depression}) generally outperform unimodal
baselines. Existing multimodal systems,
however, target adults, assume high-resource languages, and rely on
learned fusion whose parameters resist clinical audit. To our
knowledge, no prior system combines child-appropriate modalities for a
Bengali-language, Bangladesh-specific screening context.

\subsection{Explainable AI in Clinical Decision Support}
Additive feature-attribution methods such as SHAP~\cite{lundberg2017shap}
and LIME~\cite{ribeiro2016lime} dominate post-hoc explanation, and
studies of clinician needs emphasize that explanations must map onto
actionable, domain-meaningful factors rather than raw
features~\cite{tonekaboni2019clinicians}; explainability is increasingly
framed as a precondition for the responsible clinical use of
AI~\cite{amann2020explainability}. Rudin~\cite{rudin2019stop} further
argues that high-stakes decisions call for models that are
interpretable \emph{by design} rather than post-hoc explanations of
black boxes. Our training-free, clinically weighted fusion follows
that principle. Prior clinical XAI work assumes English-language reporting and
established referral infrastructure; none produces bilingual
(Bangla/English) explanations routed to a specific national
child-protection pathway.

\medskip
\noindent\textbf{Gap.} No existing work combines (a) child-appropriate
multimodal screening, (b) training-free, clinically auditable fusion,
(c) Bengali-language explainable reporting, and (d) referral routing
grounded in Bangladesh's child-protection institutions (OCC, DSS, NMHH;
Children Act 2013~\cite{childrenact2013}). ShishuRaksha AI is designed
to fill this gap as a decision-support framework, with the
synthetic-benchmark methodology of Section~IV providing a feasibility
evaluation in the absence of ethically obtainable clinical data.

\section{System Design}
\begin{figure}[t]
  \centering
  \includegraphics[width=\columnwidth]{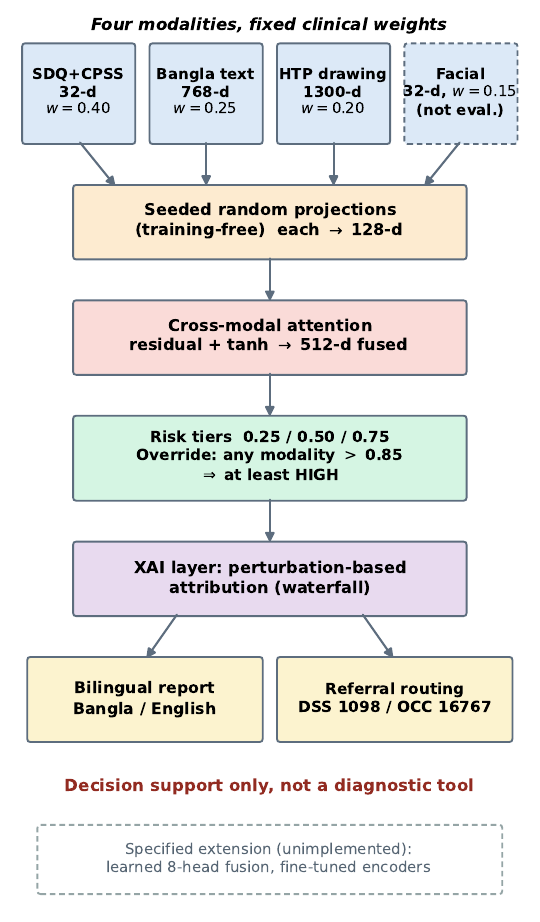}
  \caption{ShishuRaksha AI architecture: four modalities, seeded
  random projections, training-free cross-modal attention (512-d),
  tiered risk with single-modality override, and bilingual XAI reports
  with statutory referral routing.}
  \label{fig:arch}
\end{figure}

\subsection{Modalities and Clinical Weights}
Four input channels feed the system. Validated questionnaires carry
the largest clinical weight, $w_q{=}0.40$: the
SDQ~\cite{goodman1997sdq} and a DSM-5-adapted
CPSS~\cite{foa2001cpss}, with subscale scores mapped into a 32-slot
feature vector. They sit highest because they are the best-tested
instruments of the four. A
Bengali free-text narrative, encoded with
BanglaBERT~\cite{bhattacharjee2022banglabert} (768-d CLS embedding),
carries $w_t{=}0.25$. The HTP drawing channel
($w_d{=}0.20$) concatenates 1{,}280 EfficientNet-B0
features~\cite{tan2019efficientnet} with 20 binary HTP markers (omitted
figures, dark shading, heavy line pressure, encapsulation, aggressive
imagery, and similar literature-derived indicators). Facial affect is
the fourth, weakest-weighted channel ($w_f{=}0.15$), specified for
interview settings but excluded from this paper's evaluation. Fixing
these weights costs accuracy, potentially a great deal of it. What it
buys is auditability: a psychologist can read the weights, disagree
with them, and change them. No learned attention matrix allows that.

\subsection{Training-Free Attention Fusion}
Each modality vector (32-d questionnaire, 768-d text, 1{,}300-d
drawing, 32-d facial) passes through a seeded random projection into a
shared 128-d space; the projection matrices are generated from
per-modality deterministic seeds, so the entire mapping is reproducible
from the codebase alone. Scaled dot-product attention is then computed
across the four projected representations, a residual connection and
tanh nonlinearity are applied, and the four attended 128-d vectors are
concatenated into the 512-d fused representation. No parameters are
learned anywhere in this path; every weight is either clinically
specified or seed-derived, which makes the fusion fully auditable. A
learned variant (fine-tuned encoders, 8-head fusion) is specified in
the codebase but neither implemented nor evaluated here.

\subsection{Risk Stratification and Override Rule}
The fused score maps to four tiers: LOW $[0,0.25)$, MODERATE
$[0.25,0.50)$, HIGH $[0.50,0.75)$, and CRITICAL $[0.75,1]$. Each tier
binds to concrete actions and a referral target drawn from Bangladesh's
existing infrastructure: MODERATE routes to a school counsellor, HIGH
to the DSS national child helpline (1098) within 24 hours, CRITICAL to
a One-Stop Crisis Centre (hotline 16767) immediately. One safeguard sits
above the fusion: if any single modality's score exceeds 0.85, the case
is raised to at least HIGH regardless of the composite. A frightening
drawing should not be averaged away by a calm questionnaire. Whether
0.85 is the right trigger is an open clinical question; the mechanism
is the point.

\subsection{Explainability Layer}
Every risk score is decomposed by clinically weighted,
perturbation-based additive attribution in the spirit of
SHAP~\cite{lundberg2017shap}, rendered as a waterfall: which modality
pushed the score up, which pulled it down, and by how much. The choice
of a perturbation scheme over exact Shapley computation trades
theoretical guarantees for speed on commodity hardware, a trade we
consider acceptable at the screening tier; a worked example appears in
Fig.~\ref{fig:waterfall}.

\subsection{Bilingual Reporting and Referral Routing}
The output is not a number. It is a report generated in Bangla and
English side by side, stating the risk tier, the attribution waterfall
in plain language, the tier-specific action list, and the exact
referral contact with hotline number (1098 for DSS, 16767 for OCC),
consistent with the reporting duties set by the Children Act
2013~\cite{childrenact2013}. The report ends with a fixed
disclaimer, in both languages, that the tool does not diagnose.

\section{Synthetic Benchmark}
\label{sec:benchmark}
\subsection{Generation Protocol}
Collecting drawings and narratives from abused children to validate an
unproven prototype is not an option we considered. The benchmark is
generated instead, from a single fixed seed (42). Questionnaire scores
are sampled per class from SDQ and CPSS subscale distributions
consistent with published cutoffs (SDQ total difficulties 0--40, CPSS
DSM-5 subscales); drawing cases are 20-dimensional binary HTP marker
vectors with class-conditional marker prevalences taken from the HTP
literature, plus a continuous marker-burden score; Bengali narratives
are template-generated in trauma and non-trauma variants.

\subsection{Noise-Aware Design}
Clean synthetic data would prove nothing, so the generator deliberately
corrupts its own output through four layers. Questionnaire scores
receive $\pm$15\% measurement error, 10\% reporter misclassification,
and 5\% missing subscales. Drawing markers suffer 20\% detection
errors with $\pm$0.1 jitter on the burden score. Narratives flip
trauma$\to$non-trauma in 15\% of cases (modeling partial disclosure)
and the reverse in 10\%. Labels themselves carry 8\% inter-rater
disagreement flips. Each layer models a failure mode we expect in the
field; whether the four together approximate real-world messiness is
untestable until real data exist.

\subsection{Cohort Composition}
Labels are drawn at a 20\% target prevalence and then subjected to the
8\% inter-rater noise layer, yielding a realized cohort of 500 cases
with 116 positives (23.2\%). \emph{No real children and no clinical
data were used at any stage.} One circularity must be disclosed
plainly: three of the five text features used in evaluation
(emotional density, trauma keywords, disclosure score) are constructed
as noisy label-correlated proxies rather than extracted from the
generated narratives, so the text channel's measured contribution is
optimistic by construction. Results involving that modality should be
read with this in mind.

\section{Evaluation Setup}
The training-free fusion engine produces a fixed score, not a fitted
probability, so discrimination is evaluated through tree-ensemble
surrogates over the same modality features. The fused
configuration and the text-only configuration use gradient-boosted
classifiers~\cite{friedman2001gbm} (200 and default estimators,
respectively); the SDQ-only and drawing-only baselines use random
forests (100 trees); all use median imputation, a fixed random state of
42, and the fused configuration additionally receives the
clinically-weighted composite score as a feature. Stratified 5-fold
cross-validation over the 500 cases yields out-of-fold predictions for
every case, and all reported metrics are computed on the pooled
out-of-fold predictions. There is no held-out test set. With 500 cases,
we judged that setting one aside would add more statistical noise than
rigor, but readers should weigh the cross-validated numbers with that
in mind. Confidence intervals are 1{,}000-resample
bootstraps. The facial modality is excluded from evaluation entirely:
no facial component was implemented, and we prefer reporting nothing
over reporting a placeholder.

\section{Results}
\begin{table}[t]
\caption{Screening performance under 5-fold stratified CV (synthetic benchmark, $n{=}500$).}
\label{tab:main}
\centering
\begin{tabular}{lcc}
\toprule
Model & AUC & 95\% CI \\
\midrule
SDQ only (baseline) & 0.756 & [0.705, 0.803] \\
Text only & 0.768 & [0.716, 0.815] \\
Drawing only & 0.774 & [0.718, 0.827] \\
ShishuRaksha AI (fusion) & \textbf{0.874} & [0.834, 0.908] \\
\bottomrule
\end{tabular}
\end{table}

\begin{figure}[t]
  \centering
  \includegraphics[width=\columnwidth]{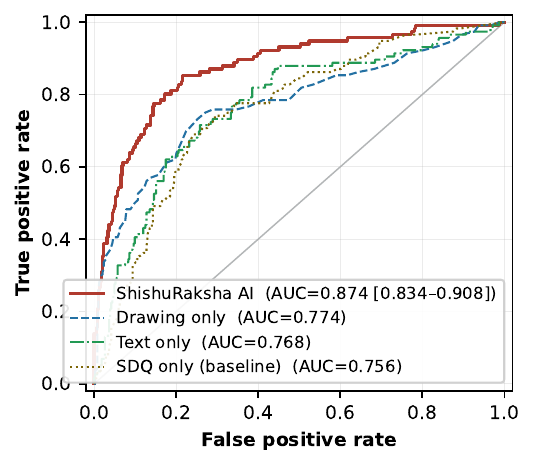}
  \caption{ROC curves under 5-fold stratified CV on the synthetic
  benchmark ($n{=}500$); all AUCs match the released evaluation
  pipeline exactly.}
  \label{fig:roc}
\end{figure}

\subsection{Comparison and Ablation}
Fig.~\ref{fig:roc} and Table~\ref{tab:main} tell a consistent story.
The three unimodal configurations land close together, between 0.756
and 0.774, and none clearly beats the SDQ-only baseline. Fusion changes
that: 0.874, with a confidence interval whose lower bound sits above
every unimodal point estimate. On synthetic data, this pattern hints
that the modalities carry partly independent signal, which is what the
clinical-weighting design assumes. It cannot show that real children's
data would behave the same way. The fused curve stays above the others
across most of the false-positive range, with the largest margin below
FPR~$\approx$~0.3, exactly the region a screening deployment would
occupy.

\begin{figure}[t]
  \centering
  \includegraphics[width=\columnwidth]{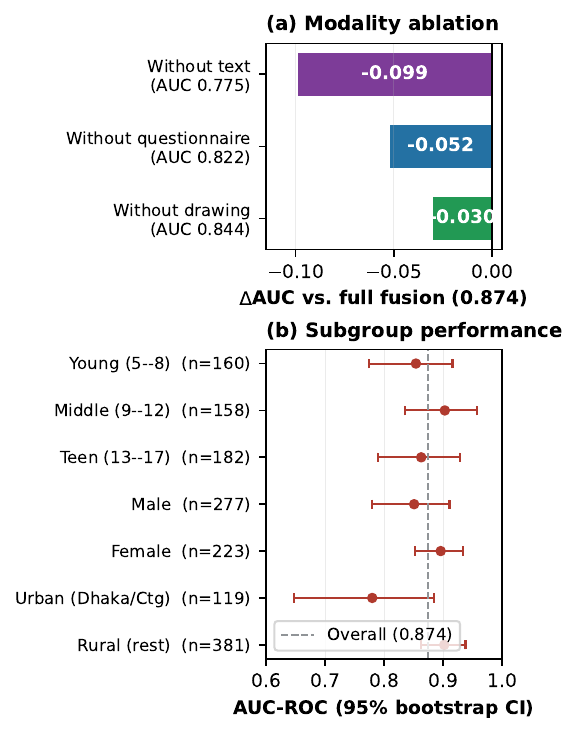}
  \caption{(a) Leave-one-modality-out ablation of the fused
  configuration. (b) Subgroup AUC-ROC with 1{,}000-resample bootstrap
  CIs; the dashed line marks the overall fused AUC. The urban subgroup
  underperforms the rural one by 0.122 AUC.}
  \label{fig:abl}
\end{figure}

Ablation (Fig.~\ref{fig:abl}a) removes one modality at a time from
the fused configuration. Dropping the text channel costs the most
($-$0.099, to 0.775). Removing the questionnaire costs $-$0.052, and
removing the drawing channel only $-$0.030. We are careful not to
celebrate the text result. It is exactly the modality carrying the
label-proxy circularity disclosed in Section~\ref{sec:benchmark}, so
its large ablation drop is partly an artifact of how its features were
built. The honest reading is that the questionnaire and drawing
channels each add modest, plausibly real signal, while the text
channel's true value will only be known once features are extracted
from real narratives.

A weight-sensitivity check adds robustness and a caveat at once.
Re-running the fused evaluation with equal weights, swapped
questionnaire and text weights, a drawing-heavy setting, and with the
composite feature removed entirely yields AUCs of 0.865--0.875. The
headline number does not depend on the chosen clinical weights; it
also means the weights add little measured discrimination, and their
value lies in the auditability of the deployed engine, not here.

\subsection{Operating Points}
At the MODERATE tier boundary (0.25), the fused model reaches
sensitivity 0.707 at specificity 0.878 (PPV 0.636, NPV 0.908). This is
the setting a broad screening pass would use: it accepts more false
alarms in order to miss fewer children. The HIGH boundary (0.50) trades down to
sensitivity 0.603 at specificity 0.930 (PPV 0.722), and the CRITICAL
boundary (0.75) reaches specificity 0.956 at sensitivity 0.457. With
only 116 positive cases, per-cell counts behind these estimates are
small and threshold-level numbers should be treated as unstable; the
same caveat applies to the four-tier confusion matrix, which is
additionally degenerate because the synthetic ground truth is binary
(LOW/HIGH only) while predictions span all four tiers.

\subsection{Subgroup Analysis}
Fig.~\ref{fig:abl}b shows fused-model AUC across age, gender, and
location subgroups. Age and gender bands stay within roughly 0.05 of
the overall estimate with overlapping intervals. Location does not:
urban cases (Dhaka/Chittagong divisions, $n{=}119$) score 0.780
[0.648--0.885] against 0.902 [0.863--0.938] for the rest of the
country. The intervals overlap, so with this sample size the gap is suggestive
rather than established. Still, a 0.122 point difference in the
direction of \emph{worse} urban performance is exactly the kind of gap
that would mean unequal screening quality in deployment, and we flag
it as a first-order item for any real-data validation.

\begin{figure}[t]
  \centering
  \includegraphics[width=\columnwidth]{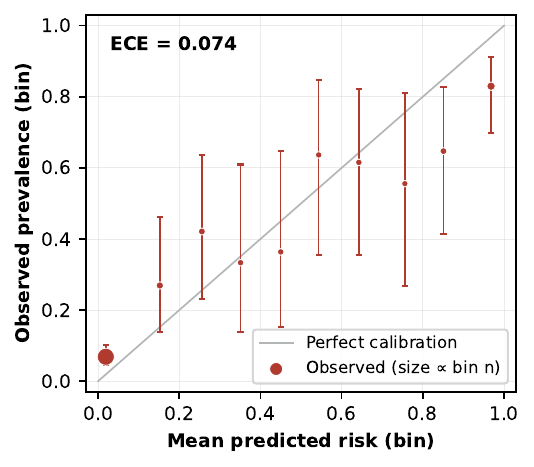}
  \caption{Reliability diagram (10 bins, out-of-fold predictions).
  Point size is proportional to bin count; error bars are 95\% Wilson
  intervals. Expected calibration error (ECE) is 0.074, with
  overconfidence visible in the highest-risk bins.}
  \label{fig:calibration}
\end{figure}

\subsection{Calibration}
Discrimination is only half the requirement; a score that overstates
risk weakens screener trust. Fig.~\ref{fig:calibration} shows
acceptable global calibration (ECE~$=0.074$) with one consistent
failure: in the highest bins the model predicts risk near 0.85--0.97
while observed prevalence sits near 0.65--0.83. Raw scores must
therefore not be read as probabilities; thresholds should be set on
ranked scores, or the output recalibrated before any probability is
shown to a human. Mid-range bins hold only 9--26 cases each, so we
avoid reading meaning into them.
\subsection{Qualitative XAI Example}
\begin{figure}[t]
  \centering
  \includegraphics[width=\columnwidth]{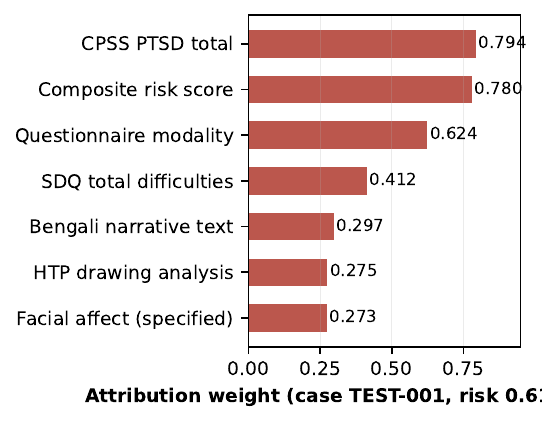}
  \caption{Attribution waterfall for synthetic case TEST-001 (risk
  0.618, tier HIGH), as rendered in the bilingual report.}
  \label{fig:waterfall}
\end{figure}

Fig.~\ref{fig:waterfall} shows the attribution output for a synthetic
test case (age 12, risk score 0.618, tier HIGH). The CPSS PTSD total
dominates, followed by the composite risk score and the questionnaire
modality as a whole; drawing and facial channels contribute least.
This ordering is what the clinical weights would predict: reassuring
for auditability, and a reminder that the explanation reflects the
design's assumptions as much as the case's evidence. The report
renders this waterfall with Bangla-first labels and the DSS 1098
referral line.

\section{Limitations and Future Work}
\label{sec:limitations}
Everything in this paper rests on synthetic data. No result says
anything direct about real Bangladeshi children; a system at AUC 0.874
on a generated cohort could sit near chance on a clinical one. The evaluation also measures tree-ensemble
surrogates rather than the training-free fusion engine itself; the
surrogates consume the same features, but the two are not the same
computation, and the gap between them is unquantified. There is no held-out test set. The
text modality inherits a circularity from the generation process, as
disclosed in Section~IV, and its contribution should be discounted
accordingly. The facial modality exists only on paper. Calibration
degrades exactly where it matters most, at high predicted risk. And
subgroup performance is uneven, with urban cases scoring 0.122 AUC
below rural ones (Fig.~\ref{fig:abl}b), a gap that would mean unequal
screening quality across communities if deployed as-is.

Some of these problems have a clear path. Prospective validation under
proper ethical review would replace the synthetic evidence or refute
it; the likely route is a partnership with a child-protection NGO and
clinical co-investigators, with assent and guardian-consent protocols
designed before any case is collected. Earlier still, Bangladeshi
child psychologists scoring a sample of synthetic cases would test
whether the benchmark's notion of risk matches clinical intuition. Recalibration, an evaluated facial channel, and
the specified deep extension all wait on data that does not yet
ethically exist. We consider that ordering correct; collecting data
first and asking ethics questions later is how this domain gets
systems that should never have been built.

\section{Conclusion}
ShishuRaksha AI is a design and a feasibility argument, not a validated
screening tool. On a noise-aware synthetic benchmark, fusing four
child-appropriate modalities under clinically specified weights
outperformed every unimodal alternative by a margin large enough to
justify the next, harder step. The framework's value may lie less in its numbers than in its
scaffolding: auditable fusion, bilingual explanation, and statutory
referral routing. That scaffolding survives even if the specific
weights do not. Whether any of this holds on real
children is precisely the question a synthetic study cannot answer,
and the one we now intend to ask properly.

\section*{Ethics, Code, and Data Availability}
This study uses exclusively synthetic data; no real children, clinical
records, or personally identifiable information were involved.
ShishuRaksha AI is designed strictly as a decision-support tool for
trained professionals and must not be used for diagnosis. The full
codebase, the synthetic benchmark, and the evaluation pipeline that
produced every number in this paper are available at
\url{https://github.com/shkoli/shishuraksha-ai}.

\balance
\bibliographystyle{IEEEtran}
\bibliography{refs}

@article{goodman1997sdq,
  author  = {Goodman, Robert},
  title   = {The Strengths and Difficulties Questionnaire: A research note},
  journal = {J. Child Psychol. Psychiatry},
  year    = {1997},
  volume  = {38},
  number  = {5},
  pages   = {581--586}
}

@article{foa2001cpss,
  author  = {Foa, Edna B. and Johnson, Kelly M. and Feeny, Norah C. and Treadwell, Kimberli R. H.},
  title   = {The Child {PTSD} Symptom Scale: A preliminary examination of its psychometric properties},
  journal = {J. Clin. Child Psychol.},
  year    = {2001},
  volume  = {30},
  number  = {3},
  pages   = {376--384}
}

@book{buck1948htp,
  author    = {Buck, John N.},
  title     = {The {House-Tree-Person} Technique},
  publisher = {Western Psychological Services},
  year      = {1948}
}

@book{malchiodi1998art,
  author    = {Malchiodi, Cathy A.},
  title     = {Understanding Children's Drawings},
  publisher = {Guilford Press},
  year      = {1998}
}

@inproceedings{bhattacharjee2022banglabert,
  author    = {Bhattacharjee, Abhik and Hasan, Tahmid and Ahmad, Wasi Uddin and Mubasshir, Kazi Samin and Islam, Md Saiful and Iqbal, Anindya and Rahman, M. Sohel and Shahriyar, Rifat},
  title     = {{BanglaBERT}: Language Model Pretraining and Benchmarks for Low-Resource Language Understanding Evaluation in {Bangla}},
  booktitle = {Findings of ACL: NAACL},
  year      = {2022},
  pages     = {1318--1327}
}

@inproceedings{tan2019efficientnet,
  author    = {Tan, Mingxing and Le, Quoc V.},
  title     = {{EfficientNet}: Rethinking Model Scaling for Convolutional Neural Networks},
  booktitle = {Proc. ICML},
  year      = {2019},
  pages     = {6105--6114}
}

@inproceedings{lundberg2017shap,
  author    = {Lundberg, Scott M. and Lee, Su-In},
  title     = {A Unified Approach to Interpreting Model Predictions},
  booktitle = {Proc. NeurIPS},
  year      = {2017},
  pages     = {4765--4774}
}

@misc{childrenact2013,
  author       = {{Government of the People's Republic of Bangladesh}},
  title        = {The Children Act, 2013 (Act No. XXIV of 2013)},
  howpublished = {Bangladesh Gazette},
  year         = {2013}
}

@article{friedman2001gbm,
  author  = {Friedman, Jerome H.},
  title   = {Greedy Function Approximation: A Gradient Boosting Machine},
  journal = {The Annals of Statistics},
  year    = {2001},
  volume  = {29},
  number  = {5},
  pages   = {1189--1232}
}

@article{goodman2001psychometric,
  author  = {Goodman, Robert},
  title   = {Psychometric Properties of the Strengths and Difficulties Questionnaire},
  journal = {J. Am. Acad. Child Adolesc. Psychiatry},
  year    = {2001},
  volume  = {40},
  number  = {11},
  pages   = {1337--1345}
}

@article{mullick2001bangla,
  author  = {Mullick, Mohammad S. I. and Goodman, Robert},
  title   = {Questionnaire Screening for Mental Health Problems in Bangladeshi Children: A Preliminary Study},
  journal = {Soc. Psychiatry Psychiatr. Epidemiol.},
  year    = {2001},
  volume  = {36},
  number  = {2},
  pages   = {94--99}
}

@article{goodman2000dhaka,
  author  = {Goodman, Robert and Renfrew, Daniel and Mullick, Mohammad},
  title   = {Predicting Type of Psychiatric Disorder from {Strengths and Difficulties Questionnaire} ({SDQ}) Scores in Child Mental Health Clinics in {London} and {Dhaka}},
  journal = {Eur. Child Adolesc. Psychiatry},
  year    = {2000},
  volume  = {9},
  number  = {2},
  pages   = {129--134}
}

@article{allen2012projective,
  author  = {Allen, Brian and Tussey, Chriscelyn},
  title   = {Can Projective Drawings Detect if a Child Experienced Sexual or Physical Abuse? {A} Systematic Review of the Controlled Research},
  journal = {Trauma, Violence, \& Abuse},
  year    = {2012},
  volume  = {13},
  number  = {2},
  pages   = {97--111}
}

@article{lin2022htp,
  author  = {Lin, Yijing and Zhang, Nan and Qu, Yukun and Li, Tian and Liu, Jia and Song, Yiying},
  title   = {The {House-Tree-Person} Test Is Not Valid for the Prediction of Mental Health: An Empirical Study Using Deep Neural Networks},
  journal = {Acta Psychologica},
  year    = {2022},
  volume  = {230},
  pages   = {103734}
}

@article{baird2022syrian,
  author  = {Baird, Sarah and Panlilio, Carlomagno and Seager, Jennifer and Smith, Sarah and Wydick, Bruce},
  title   = {Identifying Psychological Trauma Among {Syrian} Refugee Children for Early Intervention: Analyzing Digitized Drawings Using Machine Learning},
  journal = {J. Dev. Econ.},
  year    = {2022},
  volume  = {156},
  pages   = {102822}
}

@article{alshahrani2024drawing,
  author  = {Alshahrani, Amirah and Mustafa, Jumanah I. and Almatrafi, Maram M. and Aljabri, Renad A.},
  title   = {A Children's Psychological and Mental Health Detection Model by Drawing Analysis Based on Computer Vision and Deep Learning},
  journal = {Eng. Technol. Appl. Sci. Res.},
  year    = {2024},
  volume  = {14},
  number  = {4},
  pages   = {15533--15540}
}

@article{shatte2019ml,
  author  = {Shatte, Adrian B. R. and Hutchinson, Delyse M. and Teague, Samantha J.},
  title   = {Machine Learning in Mental Health: A Scoping Review of Methods and Applications},
  journal = {Psychological Medicine},
  year    = {2019},
  volume  = {49},
  number  = {9},
  pages   = {1426--1448}
}

@inproceedings{alhanai2018depression,
  author    = {Al Hanai, Tuka and Ghassemi, Mohammad M. and Glass, James},
  title     = {Detecting Depression with Audio/Text Sequence Modeling of Interviews},
  booktitle = {Proc.\ Interspeech},
  year      = {2018},
  pages     = {1716--1720}
}

@inproceedings{ribeiro2016lime,
  author    = {Ribeiro, Marco Tulio and Singh, Sameer and Guestrin, Carlos},
  title     = {``{Why} Should {I} Trust You?'': Explaining the Predictions of Any Classifier},
  booktitle = {Proc.\ 22nd ACM SIGKDD Int.\ Conf.\ on Knowledge Discovery and Data Mining},
  year      = {2016},
  pages     = {1135--1144}
}

@inproceedings{tonekaboni2019clinicians,
  author    = {Tonekaboni, Sana and Joshi, Shalmali and McCradden, Melissa D. and Goldenberg, Anna},
  title     = {What Clinicians Want: Contextualizing Explainable Machine Learning for Clinical End Use},
  booktitle = {Proc.\ Machine Learning for Healthcare Conf.\ (MLHC)},
  year      = {2019},
  pages     = {359--380}
}

@article{amann2020explainability,
  author  = {Amann, Julia and Blasimme, Alessandro and Vayena, Effy and Frey, Dietmar and Madai, Vince I.},
  title   = {Explainability for Artificial Intelligence in Healthcare: A Multidisciplinary Perspective},
  journal = {BMC Med. Inform. Decis. Mak.},
  year    = {2020},
  volume  = {20},
  pages   = {310}
}

@article{mullick2005prevalence,
  author  = {Mullick, Mohammad S. I. and Goodman, Robert},
  title   = {The Prevalence of Psychiatric Disorders Among 5--10 Year Olds in Rural, Urban and Slum Areas in Bangladesh},
  journal = {Soc. Psychiatry Psychiatr. Epidemiol.},
  year    = {2005},
  volume  = {40},
  pages   = {663--671}
}

@article{lee2024htp,
  author  = {Lee, Moonyoung and Kim, Junhyung and Kim, Jongpil},
  title   = {Generating Psychological Analysis Tables for Children's Drawings Using Deep Learning},
  journal = {Data Knowl. Eng.},
  year    = {2024},
  volume  = {149},
  pages   = {102266}
}

@article{rudin2019stop,
  author  = {Rudin, Cynthia},
  title   = {Stop Explaining Black Box Machine Learning Models for High Stakes Decisions and Use Interpretable Models Instead},
  journal = {Nature Machine Intelligence},
  year    = {2019},
  volume  = {1},
  number  = {5},
  pages   = {206--215}
}

@article{koly2024stakeholder,
  author  = {Koly, Kamrun Nahar and Saba, Jobaida and Rao, Mala and Rasheed, Sabrina and Reidpath, Daniel D. and Armstrong, Stephanie and Gnani, Shamini},
  title   = {Stakeholder Perspectives of Mental Healthcare Services in Bangladesh, Its Challenges and Opportunities: A Qualitative Study},
  journal = {Camb. Prisms Glob. Ment. Health},
  year    = {2024},
  volume  = {11},
  pages   = {e37}
}

\end{document}